\documentclass{article} 
\usepackage{iclr2026_conference,times}

\usepackage{amsmath,amsfonts,bm}

\def\eqref#1{equation~\ref{#1}}

\def\1{\bm{1}}

\DeclareMathAlphabet{\mathsfit}{\encodingdefault}{\sfdefault}{m}{sl}
\SetMathAlphabet{\mathsfit}{bold}{\encodingdefault}{\sfdefault}{bx}{n}

\usepackage{hyperref}
\usepackage{url}

\usepackage{amsmath}
\usepackage{amsfonts}
\usepackage{booktabs}
\usepackage{multirow}
\usepackage{array}          
\usepackage{amsthm}
\usepackage{graphicx} 
\usepackage{wrapfig}

\newtheorem{proposition}{Proposition}
\newcommand{\meanstd}[2]{%
  \ensuremath{#1_{\scriptscriptstyle #2}}}

\title{Learning Protein-Ligand Binding in Hyperbolic Space}

\author{Jianhui Wang$^{1,2}$\thanks{Equal contirbution.} \qquad Wenyu Zhu$^1$\footnotemark[1] \qquad Bowen Gao$^{1,3}$\footnotemark[1] \qquad  Xin Hong$^1$ \\
\textbf{Ya-Qin Zhang$^{1}$ \qquad Wei-Ying Ma$^1$ \qquad Yanyan Lan$^{1,4}$\thanks{Corresponding author.}} \\[0.25cm]
$^1$Institute for AI Industry Research (AIR), Tsinghua University, Beijing, China \\$^2$University of Electronic Science and Technology of China, Chengdu, China \\$^3$Department of Computer Science and Technology, Tsinghua University, Beijing, China \\$^4$Beijing Academy of Artificial Intelligence, Beijing, China}

\iclrfinalcopy 
\begin{document}

\maketitle

\begin{abstract}
Protein-ligand binding prediction is central to virtual screening and affinity ranking, two fundamental tasks in drug discovery. While recent retrieval-based methods embed ligands and protein pockets into Euclidean space for similarity-based search, the geometry of Euclidean embeddings often fails to capture the hierarchical structure and fine-grained affinity variations intrinsic to molecular interactions. In this work, we propose HypSeek, a hyperbolic representation learning framework that embeds ligands, protein pockets, and sequences into Lorentz-model hyperbolic space. By leveraging the exponential geometry and negative curvature of hyperbolic space, HypSeek enables expressive, affinity-sensitive embeddings that can effectively model both global activity and subtle functional differences-particularly in challenging cases such as activity cliffs, where structurally similar ligands exhibit large affinity gaps. Our mode unifies virtual screening and affinity ranking in a single framework, introducing a protein-guided three-tower architecture to enhance representational structure. HypSeek improves early enrichment in virtual screening on DUD-E from 42.63 to 51.44 (+20.7\%) and affinity ranking correlation on JACS from 0.5774 to 0.7239 (+25.4\%), demonstrating the benefits of hyperbolic geometry across both tasks and highlighting its potential as a powerful inductive bias for protein-ligand modeling. Our code is publicly available at \url{https://github.com/jianhuiwemi/HypSeek}.
\end{abstract}

\section{Introduction}
\label{introduction}
Modeling protein–ligand interactions is critical for drug discovery, where accurate binding affinity prediction underpins both large-scale virtual screening and fine-grained ligand prioritization. Virtual screening seeks to identify molecules likely to bind a given protein target from large compound libraries, often containing millions or even billions of candidates. Approaches such as molecular docking~\cite{friesner_glide_2004, trott_autodock_vina_2010} estimate binding compatibility by sampling ligand poses and scoring them with physics-based functions. While effective in small-scale settings, these methods are computationally intensive and scale poorly to modern library sizes. Unlike virtual screening, which emphasizes identifying likely binders from vast libraries, affinity ranking focuses on ordering a smaller set of candidate ligands by predicted binding strength, with physics-based techniques like free energy perturbation (FEP+)~\cite{wang2015accurate} offering high accuracy at the cost of extensive molecular dynamics simulations. These limitations restrict the practicality of traditional methods in early-stage drug discovery pipelines.

A notable shift in virtual screening came with DrugCLIP~\cite{gao2023drugclip}, which reframed the task as a dense retrieval problem. Rather than predicting binding affinity or docking poses, DrugCLIP learns contrastive embeddings of ligands and protein pockets such that interacting pairs are close in a shared Euclidean space. This design enables efficient similarity-based retrieval and allows for scalable screening across billion-scale compound libraries. Despite its promising performance and efficiency, DrugCLIP struggles to capture fine-grained interaction patterns which are essential for downstream affinity ranking. Recently, LigUnity~\cite{feng2025foundation} extends the retrieval-based framework by unifying virtual screening and affinity ranking into a single training objective. It combines contrastive learning for global interaction patterns with listwise ranking to model pocket-specific ligand preferences, aiming to jointly learn both binding likelihood and relative affinity within a unified embedding space.

While retrieval-based methods have shown strong potential, they typically embed ligands and protein pockets into Euclidean space, where distances grow linearly and the geometry does not explicitly encourage separation based on functional or activity-related differences. As a result, standard Euclidean training objectives may fail to emphasize fine-grained distinctions in binding strength, especially when molecular structures are similar.

\begin{wrapfigure}{t}{0.53\textwidth}
  \begin{center}
    \includegraphics[width=0.53\textwidth]{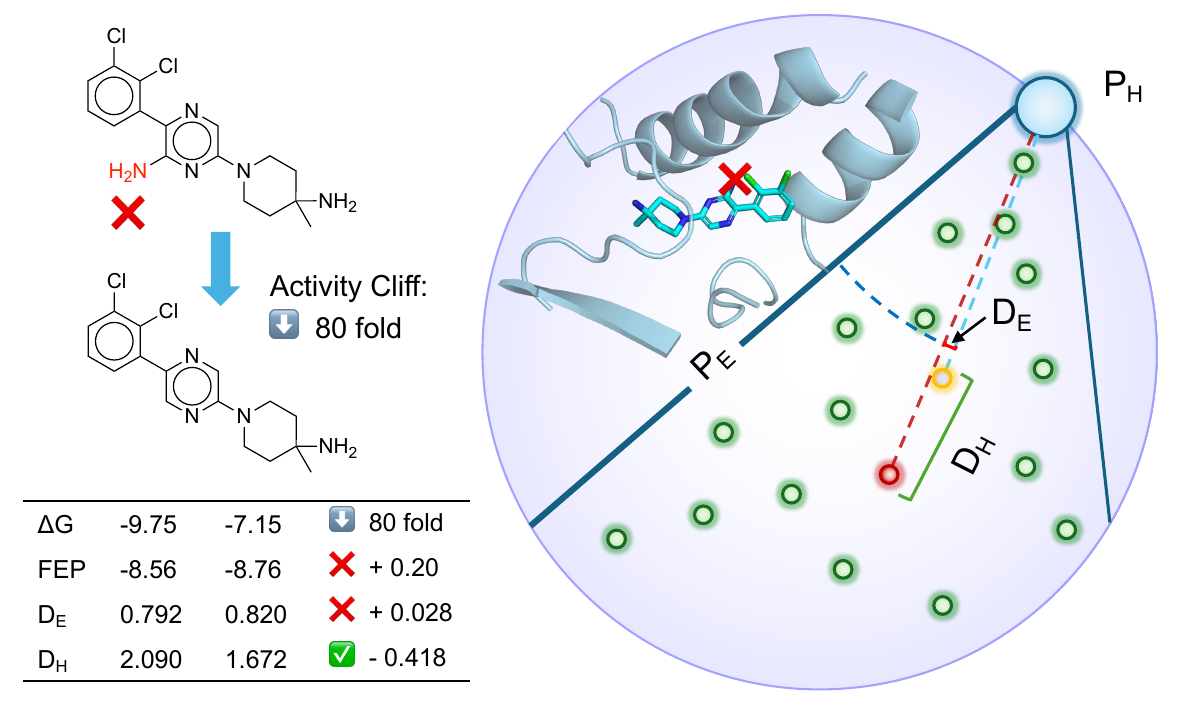}
  \end{center}
  \caption{Illustration of how hyperbolic geometry distinguishes activity cliffs (PDB ID: 5EHR). \textbf{Left}: Two structurally similar ligands (Ligand ID: 5OD vs.\ its amino-substituent-removed derivative) show an $\sim$80-fold affinity difference. \textbf{Right}: The yellow and red points denote the two ligands; the blue point is the pocket. Dashed lines show distances in hyperbolic (red/light blue) and Euclidean (dark blue) space. Euclidean embeddings preserve structural similarity but fail to reflect affinity gaps, while hyperbolic embeddings separate such pairs via both radial and angular dimensions ($D_H$, green), enabling affinity-sensitive representations.}
  \label{fig:fig1}
\end{wrapfigure}

To enrich the embedding geometry and better capture complex protein–ligand interactions, we propose HypSeek, a retrieval-based model that embeds ligands, pockets, and protein sequences into hyperbolic space. Unlike previous dual-tower designs, HypSeek adopts a protein-guided three-tower architecture during training to promote more structured representations. The curvature of hyperbolic space enables affinity-sensitive encoding through both angular direction and radial depth, providing greater expressivity than linear Euclidean geometry. This design not only enhances fine-grained affinity discrimination, but also offers a natural mechanism to address activity cliffs—cases where structurally similar ligands exhibit large differences in binding strength. While Euclidean embeddings often enforce functional similarity among structurally similar ligands, hyperbolic geometry allows such ligands to diverge meaningfully in the embedding space, reflecting differences in interaction modes or physicochemical properties. During inference, we retain efficient similarity computation via Euclidean inner products over hyperbolically shaped representations, preserving scalability without sacrificing expressiveness.

We evaluate HypSeek across both large-scale virtual screening and fine-grained affinity ranking tasks. On the DUD-E~\cite{mysinger2012dude} benchmark, HypSeek improves EF\textsubscript{1\%} from \textbf{42.63 to 51.44 (+20.7\%)}, demonstrating strong retrieval performance across targets. For affinity ranking, it increases Spearman correlation on the JACS~\cite{wang2015accurate} dataset from \textbf{0.5774 to 0.7239 (+25.4\%)}, consistently outperforming Euclidean baselines. These results highlight the benefits of hyperbolic geometry in capturing both global activity and nuanced affinity variation within a unified embedding space.

In summary, our contributions are as follows:

\begin{itemize}
\item We propose a hyperbolic embedding framework for protein–ligand modeling, where the geometry naturally captures hierarchical interactions and targets \textbf{the critical challenge of activity cliffs} by enabling structured separation of similar ligands with divergent affinities.

\item We introduce \textbf{HypSeek}, a dense retrieval model with a protein-guided three-tower architecture that integrates structure and sequence information to learn affinity-aware representations in hyperbolic space.

\item HypSeek achieves strong performance on both virtual screening and affinity ranking, capturing fine-grained binding differences more effectively than Euclidean baselines while maintaining scalable inference.
\end{itemize}

\section{Related Work}
\noindent\textbf{Virtual Screening.} Structure‐based virtual screening traditionally relies on molecular docking methods such as Glide~\cite{friesner_glide_2004} and AutoDock~\cite{trott_autodock_vina_2010}, which predict ligand binding poses and evaluate affinities using physics‐based scoring functions. Some predict binding affinity directly from protein–ligand complex structures by learning scoring functions~\cite{mcnutt_gnina_2021,shen_rtmscore_2022,cao_equiscore_2024}, while others infer interactions from raw structural inputs~\cite{lu_tankbind_2022,zhang_karmadock_2023}. A major shift occurred with DrugCLIP~\cite{gao2023drugclip}, which introduced contrastive retrieval by aligning ligand and pocket embeddings in a shared Euclidean space for billion‐scale similarity search.  This paradigm has since inspired a range of efficient retrieval methods.  For example, DrugHash~\cite{HanHongLi2025} employs binary hash codes for efficient retrieval with reduced memory cost, and LigUnity~\cite{feng2025foundation} integrates listwise ranking with contrastive screening.

\noindent\textbf{Affinity Ranking.} Accurate ranking of ligand binding affinities is essential for lead optimization but remains computationally challenging.  Physics‐based methods such as FEP+~\citep{wang2015accurate} and MM‐GB/SA~\citep{genheden2015mm} deliver high accuracy via alchemical free‐energy calculations and implicit solvent models, respectively, yet they require extensive molecular dynamics sampling.  Recent deep learning approaches seek to reduce this cost: PBCNet~\citep{yu2023computing} models pairwise ligand differences with graph neural networks, EHIGN~\citep{yang2024interaction} encodes heterogeneous protein–ligand interaction graphs, and LigUnity~\citep{feng2025foundation} combines contrastive screening with listwise ranking to jointly address global retrieval and local prioritization.

\noindent\textbf{Hyperbolic Representation Learning.} Hyperbolic space has emerged as a powerful embedding manifold for data with latent hierarchical or tree-like structure, owing to its exponential volume growth that preserves hierarchy with low distortion~\cite{nickel2017poincare, chamberlain2017neural}. Early works demonstrated that embedding taxonomies or graphs in Poincaré or Lorentz models captures hierarchical relations more faithfully than Euclidean counterparts~\cite{ganea2018hyperbolic, becigneul2018riemannian}. This theoretical appeal led to specialized optimization methods and the design of hyperbolic neural layers, including Riemannian gradient algorithms~\cite{Bonnabel13, BecigneulG19} and Hyperbolic Neural Networks~\cite{shimizu2021hyperbolicneuralnetworks}, as well as adaptations of convolutional, attention, and graph architectures~\cite{gulcehre2018hyperbolicattentionnetworks, bdeir2024fully}.  Hyperbolic embeddings have demonstrated strong performance across diverse modalities—knowledge graphs and recommender systems~\cite{liu2019hyperbolicgraphneuralnetworks, wang2021fully}, vision tasks~\cite{MettesGhadimiAtighKellerRessel2024} such as classification and few-shot learning~\cite{khrulkov2020hyperbolic, franco2023hyperbolic, liu2025hyperboliccategorydiscovery}, and language modeling~\cite{dhingra2018embedding, TifreaBG19}.  Recent studies further explore multimodal training in hyperbolic space for vision–language models to capture hierarchical semantics~\cite{desai2023hyperbolic, pal2024compositional, poppi2025hyperbolic}. Our work is the first to bring hyperbolic space to protein–ligand retrieval, leveraging its inductive bias to separate fine‐grained affinity differences.

\section{Preliminaries}
\label{sec:preliminaries}

We perform all representation learning in an $n$-dimensional hyperbolic space of constant negative curvature, using the Lorentz model~\cite{nickel2018learning,lin2023hyperbolic,desai2023hyperbolic}. This choice affords numerical stability and readily supports geodesic and exponential‐map operations.

Let $\mathbb{L}^n$ denote the Lorentz (hyperboloid) model, realized as the upper sheet
of a two‐sheeted hyperboloid in $\mathbb{R}^{n+1}$.  We first equip
$\mathbb{R}^{n+1}$ with the Lorentzian inner product
\begin{equation}
  \langle \mathbf p, \mathbf q\rangle_{\mathbb L}
  = -\,p_0\,q_0 + \bigl\langle \tilde{\mathbf p}, \tilde{\mathbf q}\bigr\rangle_{\mathbb E},
  \label{eq:lorentz-inner}
\end{equation}
where we write $\mathbf p=(p_0,\tilde{\mathbf p}),p_0\in\mathbb R,\tilde{\mathbf p}\in\mathbb R^n$
with $p_0$ the \emph{time}‐coordinate and $\tilde{\mathbf p}$ the \emph{spatial}‐coordinates, and $\langle\cdot,\cdot\rangle_{\mathbb E}$ denotes the standard Euclidean inner product.

The Lorentz model is then defined by
\begin{equation}
    \mathbb{L}^n = \Big\{ \mathbf{p} \in \mathbb{R}^{n+1} : \langle \mathbf{p}, \mathbf{p} \rangle_{\mathbb{L}} = -\frac{1}{\kappa}, p_{0} = \sqrt{\tfrac{1}{\kappa} + \|\tilde{\mathbf p}\|^{2}}, \kappa > 0 \Big\},
\end{equation}
where $-\kappa\in\mathbb{R}$ is the curvature of the space. 

We can measure distances by integrating the metric along geodesics. The Riemannian metric induced by the Lorentzian inner product gives the length of geodesics on $\mathbb L^n$, which in turn defines the hyperbolic distance.
\begin{equation}
  d_{\mathbb{L}}(\mathbf p,\mathbf q)
  = \frac1{\sqrt\kappa}\,\cosh^{-1}\!\bigl(-\,\kappa\,\langle \mathbf p,\mathbf q\rangle_{\mathbb L}\bigr),\quad \mathbf{p}, \mathbf{q} \in \mathbb{L}^n.
\label{eq:lorentz_dist}
\end{equation}

At each point $\mathbf p\in\mathbb{L}^n$, the tangent space $T_{\mathbf p}\mathbb{L}^n$
provides a linear approximation of the manifold. 
Concretely, any tangent vector $\mathbf v\in T_{\mathbf p}\mathbb{L}^n\subset\mathbb R^{n+1}$
satisfy $\langle \mathbf p,\mathbf v\rangle_{\mathbb L} = 0$, so that
\begin{equation}
  T_{\mathbf p}\mathbb{L}^n
  = \bigl\{\mathbf v\in\mathbb R^{n+1}: 
    \langle \mathbf p,\mathbf v\rangle_{\mathbb L} = 0
  \bigr\}.
  \label{eq:tangent-space}
\end{equation}

To transfer Euclidean encoder outputs into hyperbolic space, we apply the exponential map at a base point.  For any 
\(\mathbf p\in\mathbb L^n\) and \(\mathbf v\in T_{\mathbf p}\mathbb L^n\), the exponential map is
\begin{equation}
  \exp_{\mathbf p}^{\kappa}(\mathbf v)
  = \cosh\!\bigl(\sqrt\kappa\,\|\mathbf v\|_{\mathbb L}\bigr)\,\mathbf p
    + \frac{\sinh\!\bigl(\sqrt\kappa\,\|\mathbf v\|_{\mathbb L}\bigr)}
           {\sqrt\kappa\,\|\mathbf v\|_{\mathbb L}}\;\mathbf v,
  \label{eq:exp-map}
\end{equation}
where \(\|\mathbf v\|_{\mathbb L} = \sqrt{\langle \mathbf v,\mathbf v\rangle_{\mathbb L}}\). In practice, we interpret the output of a Euclidean encoder as a vector in the tangent space at the point 
\(\mathbf0=(\frac1{\sqrt\kappa},0,\dots,0)^\top\) on the hyperboloid, and then apply the exponential map 
\(\exp_{\mathbf0}^{\kappa}\) to lift it onto \(\mathbb L^n\)~\cite{Khrulkov2020}.

\section{Method}
\subsection{Problem Setting}
Our goal is to predict the binding affinity between protein pockets and candidate ligands. The training data are organized by assay, where each assay is an experimental setup designed to evaluate ligand binding against a specific protein target. Each assay includes one protein and a subset of ligands from the full compound library that have been experimentally screened, yielding binary activity labels and optionally affinity values. Crucially, affinity values are only comparable within the same assay due to differences in experimental conditions (e.g., pH, temperature, cofactors), assay protocols (e.g., cell-based or target-based), and measurement types (e.g., IC$_{50}$, $K_d$, $K_i$). 

Therefore, the task is formulated as learning relative binding strength rankings within each assay rather than predicting absolute affinities across assays. Let $\mathcal{A}$ denote the set of assays. For each assay $A_i \in \mathcal{A}$, let $\mathcal{L}_i$ be the set of tested ligands, and $v_i(\ell)$ be the affinity value of ligand $\ell \in \mathcal{L}_i$. Each assay corresponds to a target protein, represented by both its amino acid sequence and a set of candidate pocket structures $\mathcal{P}_i$. During training, one pocket from $\mathcal{P}_i$ is sampled to represent the structure, and combined with the sequence information to encode the full target. The model is trained to embed both targets and ligands into a shared hyperbolic space, enabling retrieval of active ligands and ranking them by relative binding strengths within each assay.

\begin{figure*}[t]
  \centering
  \includegraphics[width=\textwidth]{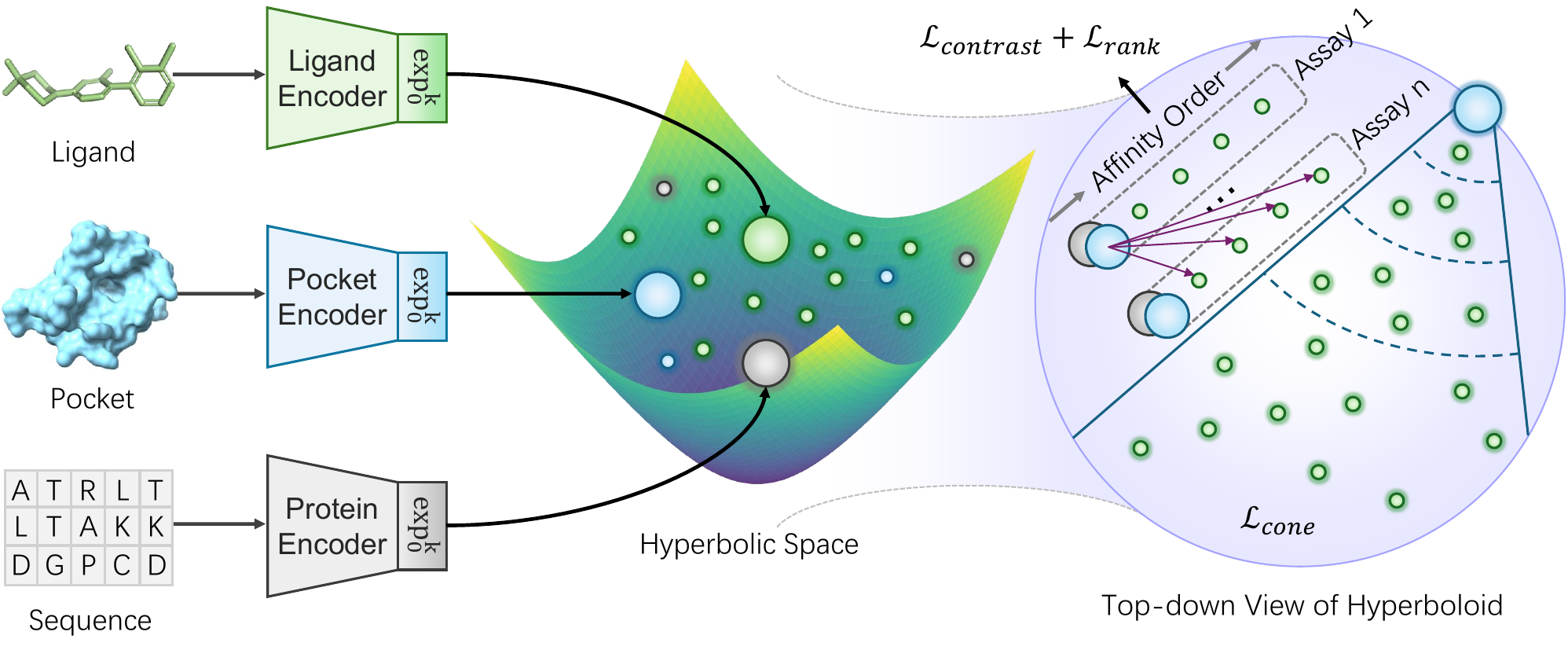}
  \caption{Overall architecture of HypSeek:  
  three encoders lift ligands, pockets and protein sequences
  to a shared hyperbolic space (left);  
  contrastive and list‑wise ranking losses align pocket/sequence with ligands
  while the cone–hierarchy loss imposes radial–angular tiers
  around each pocket (right).}
  \label{fig:framework}
\end{figure*}

\subsection{Multimodal Encoding and Lorentz Mapping}

Let \(x^{p}\) and \(x^{m}\) denote the atom-based inputs (coordinates and types) for a protein pocket and ligand, respectively, and let \(S = (s_1, \dots, s_L)\) denote the amino acid sequence of a target protein. We define three encoder functions: \(g_\phi\) and \(f_\theta\) as SE(3)-equivariant 3D graph transformers for pockets and ligands (following DrugCLIP~\citep{gao2023drugclip}), and \(h_\psi\) as a protein sequence encoder based on ESM-2~\citep{lin2023evolutionary}. As illustrated in Figure~\ref{fig:framework}, each encoder maps its input to a vector in \(\mathbb{R}^{d_{\mathrm{euc}}}\):
\begin{equation}
E_{\mathrm{poc}} = g_\phi(x^p),
E_{\mathrm{mol}} = f_\theta(x^m),
E_{\mathrm{seq}} = h_\psi(S).    
\end{equation}

We then lift these Euclidean embeddings to hyperbolic space via the exponential map defined in Eq.~\eqref{eq:exp-map}:
\begin{equation}
\begin{aligned}
\mathbf{h}_{\mathrm{poc}}
= \exp_{\mathbf 0}^{\kappa}\bigl(E_{\mathrm{poc}}\bigr),
\mathbf{h}_{\mathrm{mol}}
= \exp_{\mathbf 0}^{\kappa}\bigl(E_{\mathrm{mol}}\bigr),
\mathbf{h}_{\mathrm{seq}}
= \exp_{\mathbf 0}^{\kappa}\bigl(E_{\mathrm{seq}}\bigr).
\end{aligned}
\label{eq:encoding_and_projection}
\end{equation}

The resulting hyperbolic embeddings 
\(\mathbf{h}_{\mathrm{mol}}, \mathbf{h}_{\mathrm{poc}}, \mathbf{h}_{\mathrm{seq}} \in \mathbb{L}^n\) 
are subsequently employed in both the training and the inference stage.

\subsection{Contrastive and Ranking as the Foundation.}
We retain the in‐batch contrastive retrieval losses of DrugCLIP~\citep{gao2023drugclip} and LigUnity’s listwise ranking term~\citep{feng2025foundation}, applied to the hyperbolic embeddings
\(\tilde{\mathbf h}_u\).  For each assay \(A_i\) with query modality \(u\in\{\mathrm{poc},\mathrm{seq}\}\) and its \(B\) candidate ligands \(\{v_j\}\), we compute similarity logits $s_{i,j} \;=\; \frac{1}{\tau}\,
              \bigl\langle \tilde{\mathbf h}_{u_i},
                             \tilde{\mathbf h}_{v_j}\bigr\rangle.$
  
We adopt a symmetric InfoNCE objective over each assay \(A_i\). Let \(L_i \subseteq \{1,\dots,B\}\) denote the indices of true binders for \(u_i\). We compute:
\begin{align}
\label{eq:pl_cont}
\mathcal{L}_{\mathrm{p}\to\mathrm{l}}^{(i)}
&= -\frac{1}{|L_i|} \sum_{k \in L_i}
  \log \frac{\exp(s_{i,k})}{\sum_{j=1}^{B} \exp(s_{i,j})}, \\
\label{eq:l2p_cont}
\mathcal{L}_{\mathrm{l}\to\mathrm{p}}^{(i)}
&= -\frac{1}{|L_i|} \sum_{k \in L_i}
  \log \frac{\exp(s_{i,k})}{\sum_{n=1}^{B} \exp(s_{i,n})},
\end{align}
The total contrastive loss is then
\begin{equation}
\mathcal{L}_{\mathrm{contrast}} = \frac{1}{2} \sum_{i} \left( \mathcal{L}_{\mathrm{p}\to\mathrm{l}}^{(i)} + \mathcal{L}_{\mathrm{l}\to\mathrm{p}}^{(i)} \right).
\label{eq:total_contrast}
\end{equation}

For each assay \(A_i\) the screened ligands are sorted by measured affinity,
yielding an ordered list \((v_{i,1},\dots,v_{i,B})\).
Following the Plackett--Luce model~\cite{10.1145/1273496.1273513},
the probability of selecting ligand \(v_{i,k}\) at step \(k\)
(from the remaining set \(\mathcal R_{i,k}=\{k,k+1,\dots,B\}\))
is

\begin{equation}
\label{eq:pl_prob}
p_{i,k}(v_{i,k}) \;=\;
\frac{\exp\bigl(s_{i,k}\bigr)}
     {\displaystyle\sum_{j\in\mathcal R_{i,k}}\exp\bigl(s_{i,j}\bigr)},
\end{equation}
where \(s_{i,k}= \langle\tilde{\mathbf h}_{u_i},\tilde{\mathbf h}_{v_{i,k}}\rangle/\tau\). We use the decay $\mu_k \;=\;\frac{1}{\sqrt{B}\,\log(k+1)}$. The listwise loss for assay \(A_i\) is therefore

\begin{equation}
\label{eq:listwise_loss}
\mathcal L_{\text{rank}}^{(i)}
\;=\;
-\sum_{k=1}^{B}\mu_k
\log p_{i,k}\bigl(v_{i,k}\bigr).
\end{equation}

\subsection{Hyperbolic Geometry as a Structural Prior}
\label{sec:hyper_prior}

Beyond simply embedding pockets and ligands into a shared hyperbolic space, we aim to further leverage the geometric structure of $\mathbb{L}^n$ to encode fine-grained inductive biases about binding affinity. The exponential capacity of hyperbolic space allows for natural modeling of hierarchical relationships, while the Lorentz model enables cone-based entailment mechanisms. We therefore introduce a cone–hierarchy learning process that exploits both the radial and angular dimensions of hyperbolic space to reflect the graded nature of ligand binding strength.

Within an assay \(A_i\), the protein pocket is represented by a Lorentz‑model vector  
\(\mathbf{h}_{\mathrm{poc},i}\in\mathbb{L}^n\),  
and every screened ligand \(j\in\mathcal{L}_i\) has its own embedding  
\(\mathbf{h}_{\mathrm{mol},ij}\in\mathbb{L}^n\).  
Each hyperbolic vector splits into a time‑like coordinate and an \(n\)-dimensional spatial part: $\mathbf{h}_{\mathrm{poc},i} = \bigl(p_{0,i},\;\tilde{\mathbf{p}}_{\,i}\bigr),
\mathbf{h}_{\mathrm{mol},ij} = \bigl(m_{0,ij},\;\tilde{\mathbf{m}}_{\,ij}\bigr),$
with \(p_{0,i},m_{0,ij}\in\mathbb{R}\) and \(\tilde{\mathbf p}_{\,i},\tilde{\mathbf m}_{\,ij}\in\mathbb{R}^n\).  
These components satisfy the hyperboloid constraint  
\(p_{0,i}^{2}-\|\tilde{\mathbf p}_{\,i}\|^{2}=m_{0,ij}^{2}-\|\tilde{\mathbf m}_{\,ij}\|^{2}=1/\kappa\).  

The geodesic distance \(d_{i,j} = d_{\mathbb L}(\mathbf{h}_{\mathrm{poc},i},
\mathbf{h}_{\mathrm{mol},i,j})\) is computed via Eq.~\eqref{eq:lorentz_dist}.  
The exterior angle at the pocket,
\begin{equation}
\label{eq:phi_ij}
\phi_{i,j}
=\arccos\!\Bigl(
\frac{m_{0,i,j}
      + \kappa\bigl(\langle\tilde{\mathbf p}_i,\tilde{\mathbf m}_{i,j}\rangle
      - p_{0,i}m_{0,i,j}\bigr)p_{0,i}}
     {\|\tilde{\mathbf p}_i\|\,
      \sqrt{\,\bigl[\kappa(\langle\tilde{\mathbf p}_i,\tilde{\mathbf m}_{i,j}\rangle
      - p_{0,i}m_{0,i,j})\bigr]^2-1}} \Bigr),
\end{equation}
follows from the hyperbolic law of cosines and measures how far the ligand “leans” away from the pocket direction.

Each pocket defines a surface of admissible directions.  
Its half-aperture angle is formulated by~\citet{LeRPKN19, DesaiNR0V23} as
\begin{equation}
\label{eq:omega_i}
\omega_i
=\arcsin\!\Bigl(\tfrac{2r_0}{\sqrt{\kappa}\,\|\tilde{\mathbf p}_i\|}\Bigr),
\end{equation}
with a small constant \(r_0>0\) to keep the expression bounded near the
origin; larger \(\|\tilde{\mathbf p}_i\|\) (a pocket already pushed
towards the boundary) therefore yields a narrower cone.

Given the assay–specific affinity values
$\{v_{i,j}\}_{j=1}^{|\mathcal L_i|}$,
we draw $K$ thresholds
$t_0 < t_1 < \dots < t_K$
and assign each ligand a bucket index
\begin{equation}
  b_{i,j} \;=\; 
  \bigl\{\,k\in\{0,\dots,K\}\;:\;
         v_{i,j}\in[t_{k},t_{k+1})\,\bigr\}.    
\end{equation}
Bucket $0$ therefore collects the weakest binders and
bucket $K$ the strongest.
For every ligand we derive a bucket–specific
radial limit $r_{i,j}$ and angular–scaling factor $\eta_{i,j}$
\begin{equation}
  r_{i,j}= r_0 + b_{i,j}\,\Delta r,
  \qquad
  \eta_{i,j}= \eta_0 - b_{i,j}\,\Delta\eta,
\end{equation}
where \(r_0\) and \(\eta_0\) are the base radius/angle for the weakest tier,
and \(\Delta r,\Delta\eta>0\) are the per‑tier increments.
Smaller \(b_{i,j}\) thus yields a smaller radius cap and a larger cone. We penalise violations in
radius and angle:
\begin{align}
L_{\text{rad}}
&=\frac{1}{\sqrt{N}}\sum_{i,j}\max\bigl(d_{i,j}-r_{i,j},\,0\bigr),
\\
L_{\text{ang}}
&=\frac{1}{\sqrt{N}}\sum_{i,j}\max\bigl(\phi_{i,j}-\eta_{i,j}\,\omega_i,\,0\bigr),
\end{align}
and combine them as
\begin{equation}
\label{eq:L_chl_final}
\mathcal{L}_{\text{cone}}
=\lambda_{\text{rad}}\,L_{\text{rad}}
+\lambda_{\text{ang}}\,L_{\text{ang}}.
\end{equation}

We furthur introduce two regularization terms that operate on angular structure and intra-assay heterogeneity, respectively. To prevent trivial angular collapse, we introduce a fixed angular margin \(m > 0\) beyond the cone boundary:
\begin{equation}
R_{\text{ang}} = \frac{1}{\sqrt{N}} \sum_{i,j}
\max\bigl( \phi_{i,j} - \eta_{i,j} \omega_i + m,\; 0 \bigr),
\end{equation}

We also re-weight active ligands within each assay using rank-based weights \(w_{i,j}\) and intra-assay
softmax scores \(p_{i,j}\):
\begin{equation}
R_{\text{het}} = \frac{1}{\max(C,1)} \sum_i
\sum_{\substack{j\\v_{i,j} < v_{\mathrm{th}}}}
- w_{i,j} \log p_{i,j},
\end{equation}
where \(C\) is the number of assays with at least one active ligand,
and \(v_{\mathrm{th}}\) is a predefined affinity threshold.

\subsection{Addressing Activity Cliffs with Hyperbolic Geometry}

While structurally similar ligands often cluster in Euclidean space, such geometry can underrepresent functional differences—especially in activity cliffs, where minor structural changes lead to large affinity shifts. As formalized in Proposition~\ref{prop:hyperbolic-separation}, hyperbolic space provides exponentially greater separation via angular variation, offering a principled mechanism for distinguishing such cases. The theoretical derivation is provided in Appendix.

\begin{proposition}
\label{prop:hyperbolic-separation}
\textbf{(Hyperbolic Separation of Activity Cliffs)}  
Let $\ell_1, \ell_2$ be structurally similar ligands with large affinity differences. Under constant radial norm and small angular deviation, hyperbolic embeddings yield significantly larger geodesic distance than their Euclidean counterparts:
\[
d_{\mathbb{H}}(h_H(\ell_1), h_H(\ell_2)) \gg d_E(h_E(\ell_1), h_E(\ell_2)).
\]
This highlights the capacity of hyperbolic geometry to distinguish functionally divergent ligands without distorting local structural similarity.
\end{proposition}

\subsection{Training and Inference}
\label{sec:train_infer}
The core learning signal is driven by the pocket–ligand relationship. Accordingly, we apply hyperbolic regularisation only to the structure-based (pocket) branch, where geometric alignment in Lorentz space is both meaningful and effective. The sequence pathway provides complementary information to enhance generalisation, but does not participate in hyperbolic supervision.

Our full training objective is given by:

\begin{equation}
\begin{aligned}
\mathcal{L}_{\text{total}} \;=\;&
\underbrace{%
\alpha_{\mathrm{poc}}\!
\Bigl(
\mathcal{L}_{\text{cont}}^{\mathrm{poc}\!\leftrightarrow\!\mathrm{lig}}
+\lambda_{\text{rank}}\,
  \mathcal{L}_{\text{rank}}^{\mathrm{poc}}
\Bigr)}_{\text{pocket}\;\leftrightarrow\;\text{ligand}}
\\[4pt]
&+\,
\underbrace{%
\alpha_{\mathrm{seq}}\!
\Bigl(
\mathcal{L}_{\text{cont}}^{\mathrm{seq}\!\leftrightarrow\!\mathrm{lig}}
+\lambda_{\text{rank}}\,
  \mathcal{L}_{\text{rank}}^{\mathrm{seq}}
\Bigr)}_{\text{sequence}\;\leftrightarrow\;\text{ligand}}
\\[4pt]
&+\,
\underbrace{%
\gamma_{\text{cone}}\,\mathcal{L}_{\text{cone}}
+\lambda_{\text{ang}}\,R_{\text{ang}}
+\lambda_{\text{het}}\,R_{\text{het}}}_{\text{pocket}\;\leftrightarrow\;\text{ligand}}.
\end{aligned}
\end{equation}

At inference time, we simply embed a query pocket and each candidate ligand into hyperbolic space, extract their spatial components \(\tilde{\mathbf h}_{\mathrm{poc}}\) and \(\tilde{\mathbf h}_{\mathrm{mol},j}\), and compute similarity scores by their inner product $s_j = \tilde{\mathbf h}_{\mathrm{poc}}^\top \,\tilde{\mathbf h}_{\mathrm{mol},j}\,$.
We then rank all ligands in descending order of \(s_j\).

\section{Experiments}

\subsection{Experimental Settings}

\noindent\textbf{Implemention Details.}
We adopt the same curated assay–level training dataset as LigUnity~\cite{feng2025foundation}, which is constructed from ChEMBL~\citep{mendez_chembl_2018}, BindingDB~\cite{10.1093/nar/gkv1072}, and PDBBind~\citep{liu2017pdbbind}. For virtual screening, we strictly exclude any target UniProt IDs present in the DUD-E~\cite{mysinger2012dude}, LIT-PCBA~\cite{tran-nguyen_lit-pcba_2020} test sets. For affinity ranking tasks, we perform ligand‐level deduplication by removing redundant small molecules and non‐redundant assay IDs. Training is run on four NVIDIA A100 GPUs for 50 epochs,, using the Adam optimizer with an initial learning rate of $1\times10^{-4}$ and the curvature parameter $\kappa$ (absolute value of negative curvature) fixed to $1$.

\noindent\textbf{Benchmark.} In virtual screening, evaluations are performed on DUD-E~\cite{mysinger2012dude} and LIT-PCBA~\cite{tran-nguyen_lit-pcba_2020}. DUD-E includes 102 protein targets, each associated with experimentally verified actives and 50 property-matched decoys, designed to test enrichment capability under artificially constructed decoy scenarios. LIT-PCBA, in contrast, contains 15 targets with over 400K experimentally confirmed inactives, offering a more realistic and challenging setting without synthetic decoy bias. For affinity ranking, the evaluation is conducted on JACS~\cite{wang2015accurate} and Merck~\cite{schindler2020large}. JACS consists of eight high-quality congeneric series extracted from real lead optimization projects, emphasizing precise ranking within narrow chemical series, while Merck serves as a large-scale benchmark for FEP-based lead optimization with diverse chemical scaffolds and higher experimental noise.

\noindent\textbf{Evaluation Metrics.} For virtual screening, we use AUROC, BEDROC$_{80.5}$, Enrichment Factor (EF), and ROC-enrichment (RE) to assess model performance. For fine-grained affinity ranking, we evaluate using Pearson’s and Spearman’s rank correlation coefficients. More details are provided in Appendix~\ref{sec:eval-metrics}.

\noindent\textbf{Baselines.} We compare our method against a broad spectrum of existing approaches, including classical physics-based docking tools, empirical scoring functions, and modern deep learning models. These baselines reflect diverse modeling paradigms, ranging from structure-based simulations to neural networks trained on large protein–ligand datasets. For affinity ranking benchmarks, we additionally include methods based on free energy perturbation, energy decomposition, and recent representation learning techniques. All baselines are evaluated using their reported protocols or open-source implementations, ensuring consistency with prior work.

\subsection{Quantitative Results}

\begin{table*}[t]
    \centering
    \caption{Virtual‐screening results on the DUD-E and LIT-PCBA benchmarks.}
    \label{tab:vs_benchmarks_reduced}
    \vspace{5pt}
    \renewcommand{\arraystretch}{1.1}
    \resizebox{\textwidth}{!}{
    \begin{tabular}{l ccc c ccc}
        \toprule
        \multirow{2}{*}{Method} 
        & \multicolumn{3}{c}{\textbf{DUD-E} \;(n = 102)} && \multicolumn{3}{c}{\textbf{LIT-PCBA} \;(n = 15)}\\
        \cline{2-4} \cline{6-8}
        & AUROC & BEDROC$_{80.5}$ & EF$_{1\%}$ && AUROC & BEDROC$_{80.5}$ & EF$_{1\%}$\\
        \midrule
        Glide-SP~\cite{friesner_glide_2004}       & 0.7670 & 0.4070 & 16.18 && 0.5315 & 0.4000 & 3.41 \\
        Surflex~\cite{spitzer2012surflex}        & 0.7426 & 0.2387 & 13.35 && 0.5147 & ---    & 2.50 \\
        DeepDTA~\cite{ozturk2018deepdta}        & 0.5836 & 0.0513 &  2.28 && 0.5627 & 0.0253 & 1.47 \\
        Gnina~\cite{mcnutt2021gnina}          & 0.7817 & 0.2994 & 17.73 && 0.6093 & 0.0540 & 4.63 \\
        BigBind~\cite{brocidiacono2022bigbind}         & 0.5014 & 0.0240 &  1.18 && 0.6278 & 0.0502 & 3.79 \\
        RTMScore~\cite{shen_rtmscore_2022}       & 0.7529 & 0.4341 & 27.10 && 0.5247 & 0.0388 & 2.94 \\
        Tankbind~\cite{lu_tankbind_2022}       & 0.7509 & 0.3300 & 13.00 && 0.5970 & 0.0389 & 2.90 \\
        DrugCLIP~\cite{gao2023drugclip}       & 0.8093 & 0.5052 & 31.89 && 0.5717 & 0.0623 & 5.51 \\
        GenScore~\cite{shen2023generalized}       & 0.8160 & 0.4726 & 28.53 && 0.5957 & 0.0654 & 5.14 \\
        Planet~\cite{zhang_planet_2024}         & 0.7160 & ---    &  8.83 && 0.5731 & ---    & 3.87 \\
        EquiScore~\cite{cao_equiscore_2024}      & 0.7760 & 0.4320 & 17.68 && 0.5678 & 0.0490 & 3.51 \\
        DrugHash~\cite{Han_Hong_Li_2025}       & 0.8373 & 0.5716 & 37.18 && 0.5458 & 0.0714 & 6.14 \\
        LigUnity$_{\mathrm{poc}}$~\cite{feng2025foundation} & 0.8922 & 0.6526 & 42.63 && 0.5985 & 0.1133 & 6.47 \\
        \midrule
        \textbf{HypSeek}
                       & \textbf{0.9435} & \textbf{0.7892} & \textbf{51.44} &&
                         \textbf{0.6210} & \textbf{0.1196} & \textbf{6.81} \\
        \bottomrule
    \end{tabular}
    }
\vspace{-15pt}
\end{table*}

\noindent\textbf{Virtual Screening.} As shown in Table~\ref{tab:vs_benchmarks_reduced}, HypSeek substantially outperforms all baselines across both DUD-E and LIT-PCBA. On DUD-E, HypSeek achieves an AUROC of 0.9435, improving over the next best method (LigUnity) by more than 5 points, and delivers a BEDROC$_{80.5}$ of 0.7892, nearly 0.14 higher than LigUnity. Its EF$_{1\%}$ of 51.44 is more than 20 points above the highest competing model, demonstrating exceptional early retrieval of actives. Similarly, on the more challenging LIT-PCBA benchmark, HypSeek attains the top AUROC (0.6210), the highest BEDROC$_{80.5}$ (0.1196), and an EF$_{1\%}$ of 6.81, consistently surpassing both docking-based and deep learning approaches. These results highlight HypSeek’s superior ability to rank true binders early in the list, making it particularly well suited for high‐throughput virtual screening applications. 

\begin{wraptable}{r}{0.38\textwidth}
\centering
\vspace{-20pt} 
\caption{Complementary results.}
\vspace{5pt}
\label{complementary EF}
\resizebox{0.38\textwidth}{!}{
\renewcommand{\arraystretch}{1.1}
\begin{tabular}{l c}
\toprule
Method & EF$_{0.5\%}$ \quad EF$_{2\%}$ \quad EF$_{5\%}$ \\
\midrule
LigUnity\(_{\mathrm{poc}}\) & 48.44 \quad 29.01 \quad 13.57 \\
\textbf{HypSeek}            & \textbf{55.19 \quad 36.42 \quad 16.30} \\
\bottomrule
\end{tabular}}
\vspace{-10pt}
\end{wraptable}

We present complementary results for HypSeek on the DUD-E benchmark in Table~\ref{complementary EF}, including EF$_{0.5\%}$, EF$_{2\%}$, and EF$_{5\%}$. As shown, HypSeek outperforms LigUnity\(_{\mathrm{poc}}\) in all three EF metrics, demonstrating superior early retrieval of actives. Table~\ref{RE} reports ROC Enrichment (RE) metrics on the DUD–E benchmark under the fine‐tuning setting. Notably, HypSeek achieves RE$_{0.5\%}=137.15$, surpassing even the few-shot DrugCLIP$_\mathrm{FT}$ result of 118.10, highlighting its exceptional ability to enrich actives early in the ranking.

\begin{table}[htbp]
\centering
\caption{ROC–enrichment (RE) on the DUD–E benchmark.}
\vspace{5pt}
\label{RE}
\resizebox{\textwidth}{!}{
\renewcommand{\arraystretch}{1.12}
\begin{tabular}{l cccc c}
\toprule
Method & AUROC & RE$_{0.5\%}$ & RE$_{1\%}$ & RE$_{2\%}$ & RE$_{5\%}$ \\
\midrule
Graph CNN~\cite{doi:10.1021/acs.jcim.9b00628}             & 0.8860 & 44.41 & 29.75 & 19.41 & 10.74 \\
DrugVQA~\cite{Zheng2020}                                & 0.9720 & 88.17 & 58.71 & 35.06 & 17.39 \\
AttentionSiteDTI~\cite{10.1093/bib/bbac272}             & 0.9710 & 101.74 & 59.92 & 35.07 & 16.74 \\
COSP~\cite{Gao2023CoSP}                 & 0.9010 & 51.05 & 35.98 & 23.68 & 12.21 \\
DrugCLIP$_{\mathrm{ZS}}$~\cite{gao2023drugclip}         & 0.8093 & 73.97 & 41.79 & 23.68 & 11.16 \\
DrugCLIP$_{\mathrm{FT}}$~\cite{gao2023drugclip}         & \textbf{0.9659} & 118.10 & 67.17 & 37.17 & 16.59 \\
LigUnity\(_{\mathrm{poc}}\)~\cite{feng2025foundation}   & 0.8922 & 104.69 & 57.47 & 33.76 & 13.88 \\
\midrule
\textbf{HypSeek}                                        & 0.9435 & \textbf{137.15} & \textbf{73.16} & \textbf{38.80} & \textbf{16.60} \\
\bottomrule
\end{tabular}}
\end{table}

\noindent\textbf{Affinity Ranking.} We evaluate HypSeek on the JACS and Merck datasets using five independent random seeds to assess both accuracy and robustness. We report two sets of our results: “ensemble,” which averages the five models’ predictions before computing metrics, and “mean\(_\text{std}\),” which gives the mean and standard deviation of Pearson’s \(r\) and Spearman’s \(\rho\) across the five runs. As shown in Table~\ref{tab:affinity_benchmarks}, on JACS HypSeek (ensemble) achieves Pearson \(r=0.7742\) and Spearman \(\rho=0.7819\), closely matching the physics‐based FEP+ (Pearson \(r=0.7811\), Spearman \(\rho=0.7595\)) and significantly outperforming all deep‐learning baselines. On Merck, HypSeek (ensemble) attains Pearson \(r=0.6120\) and Spearman \(\rho=0.5447\), leading the non‐physics methods. Moreover, HypSeek’s standard deviations are lower than those reported for LigUnity’s mean\(_\text{std}\) results, indicating more consistent performance across random seeds.

\begin{table*}[htbp]
    \centering
    \caption{Affinity ranking results on the JACS and MERCK benchmark datasets.}
    \label{tab:affinity_benchmarks}
    \vspace{5pt}
    \renewcommand{\arraystretch}{1.1}
    \resizebox{\textwidth}{!}{
    \begin{tabular}{llcc c cc}
        \toprule
        \multirow{2}{*}{\textbf{Type}} & \multirow{2}{*}{\textbf{Method}} 
          & \multicolumn{2}{c}{\textbf{JACS}} 
          && \multicolumn{2}{c}{\textbf{Merck}} \\
        \cmidrule(lr){3-4} \cmidrule(lr){6-7}
         &  & Pearson~\(r\) & Spearman~\(\rho\) 
           && Pearson~\(r\) & Spearman~\(\rho\) \\
        \midrule
        \multirow{2}{*}{Physics}
         & FEP+~\cite{wang2015accurate}      & 0.7811 & 0.7595 && 0.6960 & 0.6798 \\
         & MM-GB/SA~\cite{genheden2015mm}  & 0.1489 & 0.2011 && 0.1299 & 0.1299 \\
        \midrule
        \multirow{7}{*}{DL}
         & PBCNet~\cite{yu2023computing}        & 0.3939 & 0.3799 && 0.4058 & 0.4075 \\
         & EHIGN~\cite{yang2024interaction}         & 0.5787 & 0.5814 && 0.4246 & 0.3830 \\
         & GET~\cite{pmlr-v235-kong24b}           & 0.4034 & 0.3753 && 0.4203 & 0.4214 \\
         & BindNet~\cite{feng2024proteinligand}       & 0.5481 & 0.5368 && 0.4037 & 0.3477 \\
         & Boltz-2~\cite{Passaro2025.06.14.659707}       & 0.5231 & 0.5285 && 0.4298 & 0.4013 \\
         & LigUnity$_{\mathrm{poc}}$ (ensemble)~\cite{feng2025foundation}      & 0.6454 & 0.6460 && 0.5997 & 0.5554 \\
         & LigUnity$_{\mathrm{poc}}$ (mean$_{\text{std}}$)~\cite{feng2025foundation} 
           & \meanstd{0.5705}{0.1955} & \meanstd{0.5774}{0.2097}
           && \meanstd{0.5323}{0.1865} & \meanstd{0.4994}{0.1773} \\
        \midrule
        \multirow{2}{*}{Ours}
         & \textbf{HypSeek (ensemble)} & 0.7742 & 0.7819 && 0.6120 & 0.5447 \\
         & \textbf{HypSeek (mean$_{\text{std}}$)}
           & $\mathbf{\meanstd{0.7186}{0.1157}}$ & $\mathbf{\meanstd{0.7239}{0.1321}}$
           && $\mathbf{\meanstd{0.5606}{0.1738}}$ & $\mathbf{\meanstd{0.5034}{0.1739}}$ \\
        \bottomrule
    \end{tabular}
    }
\end{table*}

\subsection{Ablation and Analysis of HypSeek}

\noindent\textbf{Impact of Key Components.} As summarised in Table~\ref{tab:combined_ablation}, switching off hyperbolic–specific terms (no hyp) already degrades virtual–screening performance on DUD-E (BEDROC$_{80.5}$ drops from 0.7892 to 0.7671; EF$_{1\%}$ from 51.44 to 49.14), while the Euclidean baseline is markedly worse. The advantage becomes even more pronounced for affinity ranking on JACS, where Pearson~$r$ falls from 0.7518 to 0.6839 without hyperbolic supervision and to 0.5978 in purely Euclidean space. In the affinity ranking task, due to limited computational resources, we conducted each ablation with a single random seed. Ablating either the angular or heterogeneity regulariser alone (no $R_{\mathrm{ang}}$, no $R_{\mathrm{het}}$) yields intermediate losses, confirming that both angle control and intra-assay weighting contribute complementary signals beyond the core cone loss. Finally, removing the protein sequence pathway (no Seq) also degrades performance, indicating that protein‐sequence features serve mainly as an auxiliary signal that further shapes the embeddings.

\begin{table*}[htbp]
    \centering
    \caption{Ablation results on the DUD-E and JACS benchmarks.}
    \vspace{5pt}
    \label{tab:combined_ablation}
    \vspace{5pt}
    \renewcommand{\arraystretch}{1.1}
    \resizebox{\textwidth}{!}{
    \begin{tabular}{l|cccc|cc|cc}
        \toprule
        \multirow{2}{*}{\textbf{Setting}}
          & \multicolumn{4}{c|}{\textbf{Module}} 
          & \multicolumn{2}{c|}{\textbf{DUD-E (n = 102)}} 
          & \multicolumn{2}{c}{\textbf{JACS}} \\
        \cmidrule(lr){2-5} \cmidrule(lr){6-7} \cmidrule(lr){8-9}
          & $\mathcal{L}_{\mathrm{cone}}$ & $R_{\mathrm{ang}}$ & $R_{\mathrm{het}}$ & Seq
          & BEDROC$_{80.5}$ & EF$_{1\%}$
          & Pearson $r$ & Spearman $\rho$ \\
        \midrule
        \multicolumn{9}{l}{\textbf{Hyperbolic space}} \\
        Full model                    & \checkmark & \checkmark & \checkmark & \checkmark & 0.7892 & 51.44 & 0.7518 & 0.7580 \\
        -- no hyp                     & $\times$   & $\times$   & $\times$   & \checkmark & 0.7671 & 49.14 & 0.6839 & 0.6906 \\
        -- no $R_{\mathrm{ang}}$      & \checkmark & $\times$   & \checkmark & \checkmark & 0.7856 & 50.52 & 0.7340     & 0.7529     \\
        -- no $R_{\mathrm{het}}$      & \checkmark & \checkmark & $\times$   & \checkmark & 0.7773 & 50.42 & 0.7047     & 0.7074     \\
        -- no Seq                     & \checkmark & \checkmark & \checkmark & $\times$   & 0.7351 & 47.70 & 0.7194 & 0.7050 \\
        \midrule
        \multicolumn{9}{l}{\textbf{Euclidean space}} \\
        Contrastive + rank   & $\times$   & $\times$   & $\times$   & $\times$   & 0.6565 & 42.87 & 0.5978 & 0.6060 \\
        \bottomrule
    \end{tabular}
    }
\end{table*}

\begin{figure*}[htbp]
  \centering
  \includegraphics[width=1.0\textwidth]{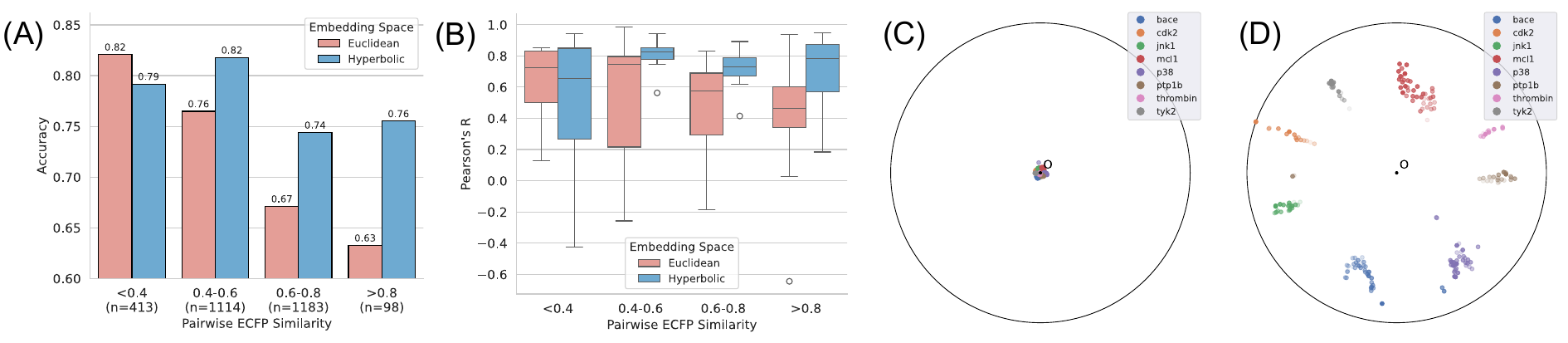}
  \caption{Pairwise analysis and CO-SNE visualization on the JACS benchmark. (A) Accuracy of affinity change prediction on ligand pairs with different ECFP4 similarity, comparing Euclidean and hyperbolic spaces; (B) Pearson's $R$ between predicted score difference and ground truth affinity gap; (C) CO-SNE visualization of ligand embeddings in hyperbolic space without the hyperbolic constraint loss; (D) CO-SNE visualization of our HypSeek ligand embeddings.}
  \label{fig:pairsim_cosne}
\end{figure*}

\noindent\textbf{Pairwise Affinity Prediction.}  
Figure~\ref{fig:pairsim_cosne} (A)-(B) demonstrate the behavior of Euclidean and hyperbolic models across varying ECFP4~\cite{doi:10.1021/ci100050t} similarity. Both models perform similarly on dissimilar ligand pairs, but as the ligands become more structurally similar, Euclidean accuracy and correlation decrease significantly. In contrast, the hyperbolic model maintains strong performance, even in these highly similar pairs. This suggests that the richer geometry information in hyperbolic space, which better accommodates relationships between molecules, is more effective at capturing subtle affinity shifts typical of situations where structurally similar molecules exhibit significantly different biological activity. These differences are often compressed in Euclidean space, where the geometry may fail to distinguish between such subtle shifts.

\noindent\textbf{Embedding Visualization.} Ligand embeddings are first reduced via HoroPCA~\cite{pmlr-v139-chami21a} and visualized using CO-SNE~\cite{9879850}. Without hyperbolic constraints (Figure~\ref{fig:pairsim_cosne}C), embeddings collapse near the origin with overlapping targets. With the full HypSeek objective (Figure~\ref{fig:pairsim_cosne}D), clear target-wise clusters and radial affinity gradients emerge. This contrast illustrates how the cone–hierarchy constraints introduced by HypSeek structure the hyperbolic manifold, enabling a more effective representation of the complex relationships between ligands in hyperbolic space.

\section{Conclusion}
We introduced HypSeek, a hyperbolic protein–ligand binding prediction model that embeds ligands, protein pockets, and sequences into a shared hyperbolic space using a three-tower architecture. By leveraging the negative curvature and exponential geometry of hyperbolic space, HypSeek captures both global interaction patterns and fine-grained affinity differences—especially in challenging cases like activity cliffs, where Euclidean embeddings often fail. Meanwhile, it retains efficient retrieval through inner product similarity, enabling large-scale virtual screening. Extensive experiments show that HypSeek consistently outperforms existing baselines across both screening and ranking tasks. HypSeek provides a geometry-aware solution for binding prediction. 
\section{Acknowledgments}
This work was supported by the National Key R\&D Program of China No. 2025ZD1802501,
the National Natural Science Foundation of China (NSFC) No. 62506193,
and the Beijing Academy of Artificial Intelligence.

\bibliography{iclr2026_conference}
\bibliographystyle{iclr2026_conference}

\clearpage
\appendix
\appendix

\section{Theoretical Motivation for Hyperbolic Separation of Activity Cliffs}
\label{theory}

A key challenge in protein--ligand modeling is the presence of \emph{activity cliffs}---cases where structurally similar ligands exhibit large differences in binding affinity. We aim to show, from a geometric perspective, why hyperbolic space is better suited than Euclidean space for separating such ligand pairs.

\subsection{Problem Setup}

Let $\ell_1, \ell_2 \in \mathbb{R}^n$ be two ligands with high structural similarity, such that their Euclidean distance is small:
\begin{equation}
\| \ell_1 - \ell_2 \|_E = \varepsilon, \quad \varepsilon \ll 1
\end{equation}
but their binding affinities differ significantly:
\begin{equation}
|f(\ell_1) - f(\ell_2)| \gg 0
\end{equation}
Our goal is to learn an embedding $h(\cdot)$ such that:
\begin{equation}
\| h(\ell_1) - h(\ell_2) \| \gg \varepsilon
\end{equation}
i.e., the embedding space should amplify functional differences despite structural similarity.

\subsection{Limitations of Euclidean Geometry}

In Euclidean space $\mathbb{R}^d$, distance grows linearly:
\begin{equation}
d_E(x, y) = \|x - y\|_2
\end{equation}
Thus, structurally similar ligands must be mapped to nearby locations unless we distort the local geometry, which harms generalization and smoothness.

\subsection{Hyperbolic Geometry and Exponential Separation}

We consider the Lorentz model of hyperbolic space $\mathbb{H}^n$ with curvature $-\kappa$. The manifold is defined as:
\begin{equation}
\mathbb{H}^n = \{ x \in \mathbb{R}^{n+1} \, | \, \langle x, x \rangle_L = -\frac{1}{\kappa}, \, x_0 > 0 \}
\end{equation}
where the Lorentzian inner product is:
\begin{equation}
\langle x, y \rangle_L = -x_0 y_0 + \sum_{i=1}^n x_i y_i
\end{equation}
The geodesic distance between $x, y \in \mathbb{H}^n$ is given by:
\begin{equation}
d_{\mathbb{H}}(x, y) = \frac{1}{\sqrt{\kappa}} \cosh^{-1} \left( -\kappa \langle x, y \rangle_L \right)
\end{equation}

\subsection{Angular Separation and Activity Cliffs}

Let $v_1, v_2 \in T_o\mathbb{H}^n$ be tangent vectors at the origin $o$, representing two structurally similar ligands. Their exponential map into $\mathbb{H}^n$ is:
\begin{equation}
\exp_o(v) = \cosh(\|v\|) \cdot o + \sinh(\|v\|) \cdot \frac{v}{\|v\|}
\end{equation}
Assume both vectors have the same norm $\|v_1\|\!\approx\!\|v_2\|\!=\!r$ (i.e.\ equal radial depth) and a small angular deviation
\(
\theta=\angle(v_1,v_2)\ll 1
\).
By applying the hyperbolic law of cosines and the expansion
\(
\operatorname{arccosh}(1+\varepsilon)=\sqrt{2\varepsilon}+{\cal O}(\varepsilon^{3/2})
\),
their geodesic distance satisfies
\begin{equation}
\label{eq:small_angle}
d_{\mathbb{H}}\!\bigl(\exp_o(v_1),\exp_o(v_2)\bigr)
\;\approx\;
\frac{\sinh r}{\sqrt{\kappa}}\;\theta
\;+\;\mathcal{O}(\theta^{3}).
\end{equation}
This implies that even small angular differences (e.g., from subtle functional changes) lead to large separations if radial depth (i.e., binding strength) differs.

\subsection{Conclusion}

\textbf{Proposition.} Let $\ell_1, \ell_2 \in \mathbb{R}^n$ be structurally similar ligands with different affinity labels. Let $h_E: \mathbb{R}^n \to \mathbb{R}^d$ be a Euclidean embedding and $h_H: \mathbb{R}^n \to \mathbb{H}^d$ a hyperbolic embedding. Then under constant radial norm $r$ and small angular separation $\theta$, we have:
\begin{equation}
d_{\mathbb{H}}(h_H(\ell_1), h_H(\ell_2)) \gg d_E(h_E(\ell_1), h_E(\ell_2))
\end{equation}
This shows that hyperbolic geometry provides stronger capacity to distinguish \emph{activity cliff pairs}, even under tight structural similarity, without requiring large Euclidean displacement or model distortion.

\section{Supplementary Analysis and Details}

\begin{table*}[t]
  \centering
  \caption{Cross-Target Activity-Cliff Cases.}
  \vspace{5pt}
  \label{tab:cliff-examples}
  \resizebox{\textwidth}{!}{%
  \begin{tabular}{c c c c c c c c c c c c}
    \toprule
    Molecule A & Molecule B & PDB ID & ECFP 
      & Exp\,$\Delta G$\,(A/B) 
      & FEP\,$\Delta G$\,(A/B) 
      & Euc\,(A/B) 
      & Hyp\,(A/B) 
      & $\Delta$(Exp\,$\Delta G$) 
      & $\Delta\text{FEP}$ 
      & $\Delta\text{Euc}$
      & $\Delta\text{Hyp}$ \\
    \bottomrule

    \raisebox{-0.5\height}{\includegraphics[width=0.10\linewidth]{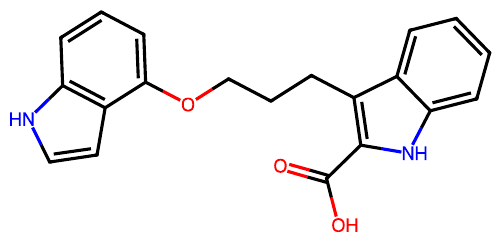}} &
    \raisebox{-0.5\height}{\includegraphics[width=0.10\linewidth]{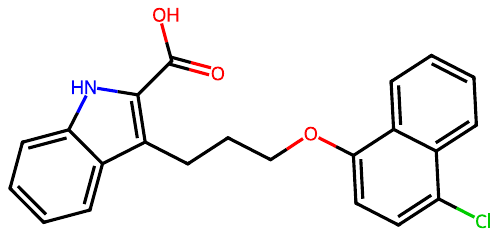}} &
    4HW3 & 
    0.6731 & 
    -6.66\,/\, -8.67 &    
    -3.5197\,/\, -8.0306 &    
    +0.5356\,/\, +0.5490 &    
    +1.5055\,/\, +1.7663 &    
    -2.01 &               
    -4.5109 &                 
    +0.0132 &               
    +0.2608 \\              

    \midrule

    \raisebox{-0.5\height}{\includegraphics[width=0.10\linewidth]{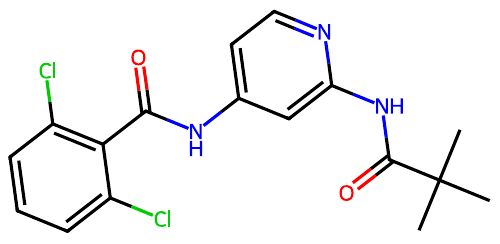}} &
    \raisebox{-0.5\height}{\includegraphics[width=0.10\linewidth]{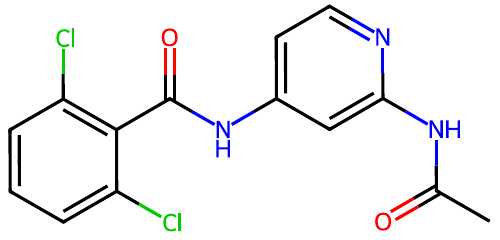}} &
    4GIH & 
    0.7674 & 
    -7.42\,/\, -9.54 & 
    -6.1068\,/\, -9.8942 &
    +0.7420\,/\, +0.7476 &
    +1.1878\,/\, +1.7734 &
    -2.12 &
    -3.7874 &               
    +0.0054 &
    +0.5855 \\

    \midrule

    \raisebox{-0.5\height}{\includegraphics[width=0.10\linewidth]{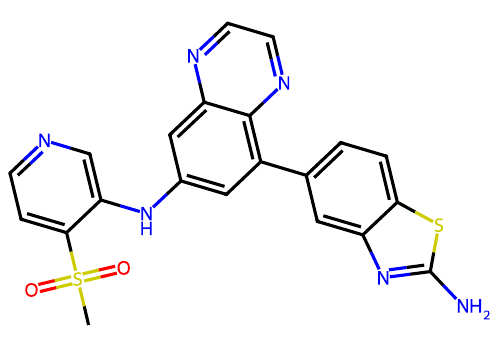}} &
    \raisebox{-0.5\height}{\includegraphics[width=0.10\linewidth]{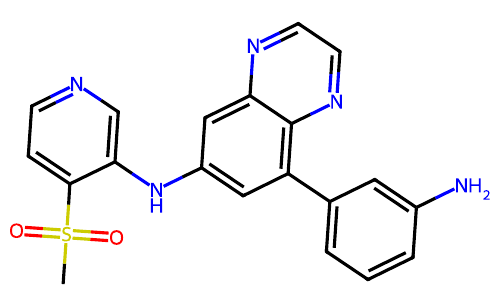}} &
    6HVI & 
    0.7097 & 
    -10.24\,/\, -7.19 & 
    -11.3100\,/\, -7.1480 &
    +0.6760\,/\, +0.6830 &
    +1.9901\,/\, +1.5151 &
    +3.05 &
    +4.16 &               
    +0.0073 &
    -0.4750 \\

    \midrule

    \raisebox{-0.5\height}{\includegraphics[width=0.10\linewidth]{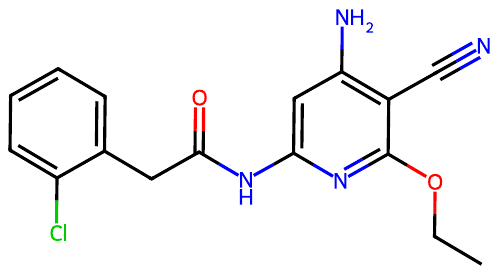}} &
    \raisebox{-0.5\height}{\includegraphics[width=0.10\linewidth]{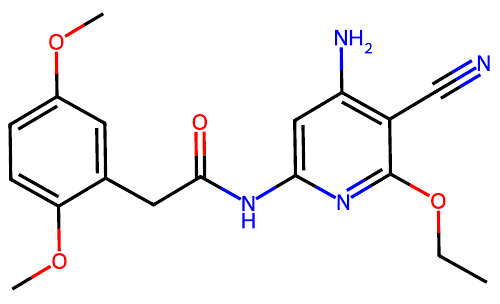}} &
    2GMX & 
    0.6500 & 
    -8.11\,/\, -9.99 & 
    -7.8979\,/\, -9.9855 &
    +0.8066\,/\, +0.7866 &
    +1.8717\,/\, +2.2653 &
    -1.88 &
    -2.0876 &               
    -0.0200 &
    +0.3936 \\

    \midrule

    \raisebox{-0.5\height}{\includegraphics[width=0.10\linewidth]{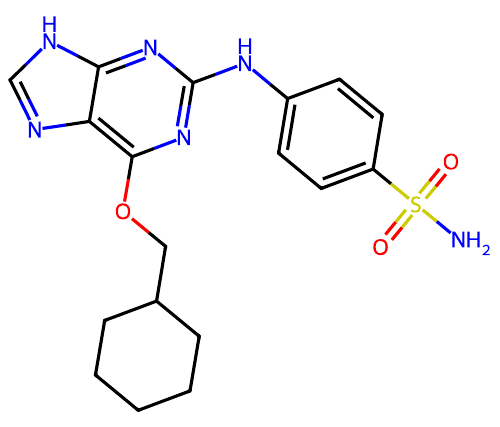}} &
    \raisebox{-0.5\height}{\includegraphics[width=0.10\linewidth]{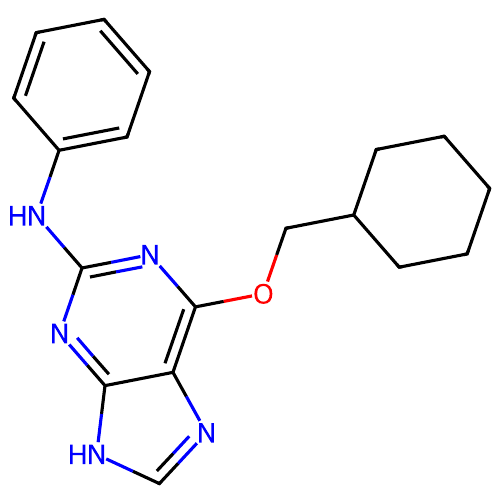}} &
    1H1Q & 
    0.7273 & 
    -11.25\,/\, -8.18 & 
    -9.8937\,/\, -8.1376 &
    +0.5684\,/\, +0.5700 &
    +2.0705\,/\, +1.7103 &
    +3.07 &
    +1.7561 &               
    +0.0015 &
    -0.3601 \\

    \midrule

    \raisebox{-0.5\height}{\includegraphics[width=0.10\linewidth]{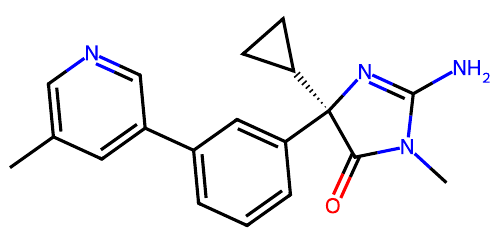}} &
    \raisebox{-0.5\height}{\includegraphics[width=0.10\linewidth]{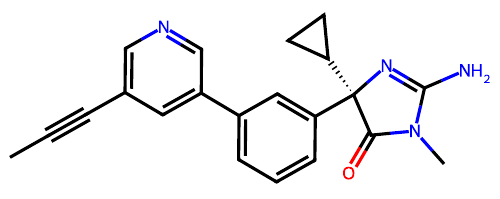}} &
    4DJW & 
    0.7719 & 
    -9.47\,/\, -11.35 & 
    -9.9357\,/\, -11.1010 &
    +0.9110\,/\, +0.9175 &
    +2.1507\,/\, +2.3914 &
    -1.88 &
    -1.1653 &               
    +0.0063 &
    +0.2407 \\

    \midrule

    \raisebox{-0.5\height}{\includegraphics[width=0.10\linewidth]{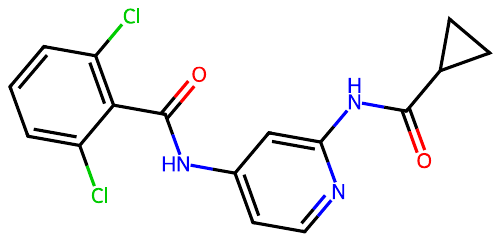}} &
    \raisebox{-0.5\height}{\includegraphics[width=0.10\linewidth]{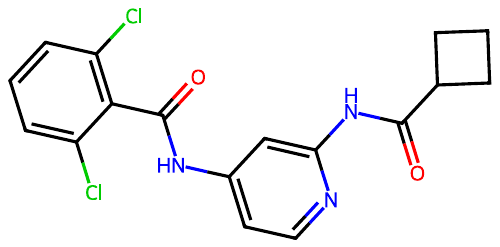}} &
    4GIH & 
    0.9048 & 
    -11.31\,/\, -9.70 & 
    -10.5581\,/\, -9.4767 &
    +0.7390\,/\, +0.7446 &
    +1.8883\,/\, +1.6953 &
    +1.61 &
    +1.0814 &               
    +0.0059 &
    -0.1930 \\

    \midrule

    \raisebox{-0.5\height}{\includegraphics[width=0.10\linewidth]{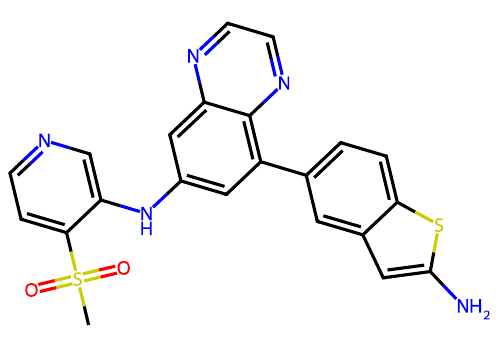}} &
    \raisebox{-0.5\height}{\includegraphics[width=0.10\linewidth]{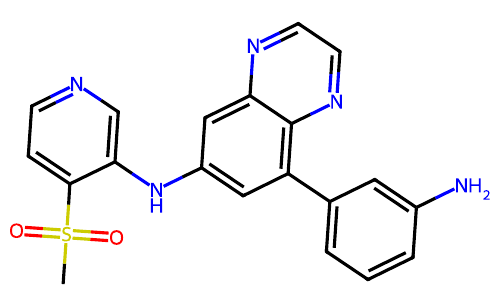}} &
    6HVI & 
    0.7213 & 
    -9.77\,/\, -7.19 & 
    -11.1920\,/\, -7.1480 &
    +0.6953\,/\, +0.6830 &
    +2.0241\,/\, +1.5151 &
    +2.58 &
    +4.0440 &               
    -0.0122 &
    -0.5090 \\

    \midrule

    \raisebox{-0.5\height}{\includegraphics[width=0.10\linewidth]{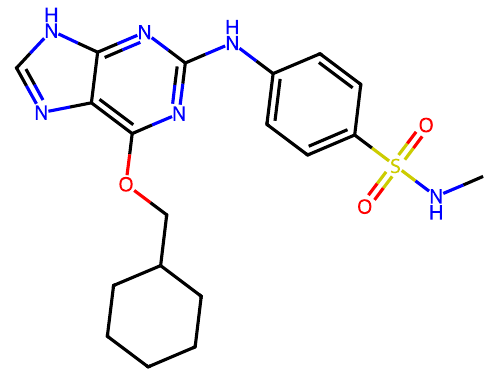}} &
    \raisebox{-0.5\height}{\includegraphics[width=0.10\linewidth]{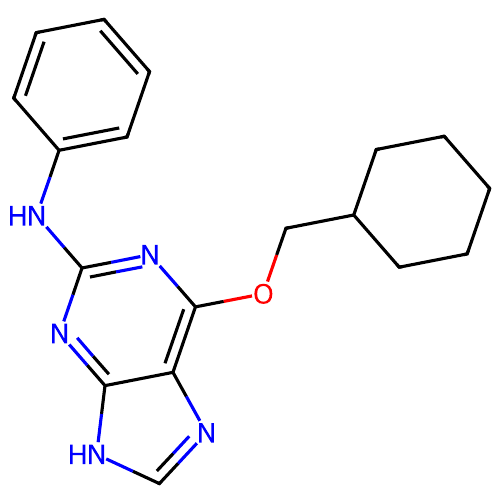}} &
    1H1Q & 
    0.7018 & 
    -11.11\,/\, -8.18 & 
    -9.8570\,/\, -8.1376 &
    +0.5645\,/\, +0.5700 &
    +2.0925\,/\, +1.7103 &
    +2.93 &
    +1.7194 &               
    +0.0054 &
    -0.3822 \\

    \midrule

    \raisebox{-0.5\height}{\includegraphics[width=0.10\linewidth]{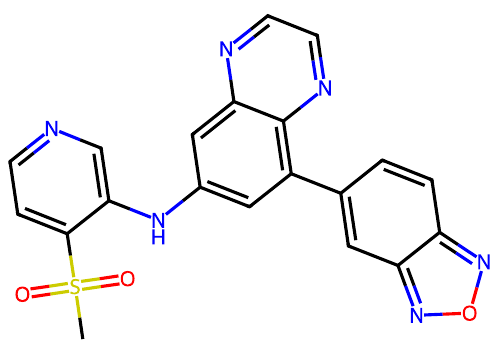}} &
    \raisebox{-0.5\height}{\includegraphics[width=0.10\linewidth]{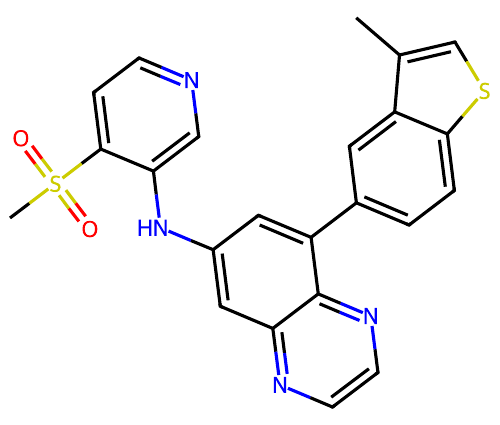}} &
    6HVI & 
    0.6515 & 
    -7.69\,/\, -10.71 & 
    -8.6460\,/\, -10.5600 &
    +0.7173\,/\, +0.7380 &
    +1.5800\,/\, +2.2676 &
    -3.02 &
    -1.9140 &               
    +0.0205 &
    +0.6876 \\

    \midrule

    \raisebox{-0.5\height}{\includegraphics[width=0.10\linewidth]{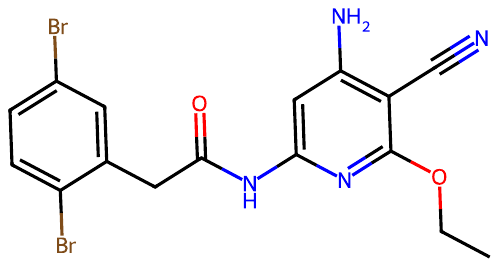}} &
    \raisebox{-0.5\height}{\includegraphics[width=0.10\linewidth]{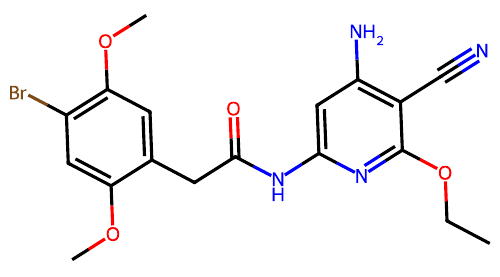}} &
    2GMX & 
    0.6897 & 
    -7.51\,/\, -9.68 & 
    -8.8421\,/\, -10.7494 &
    +0.7437\,/\, +0.7650 &
    +1.9024\,/\, +2.3753 &
    -2.17 &
    -1.9073 &               
    +0.0215 &
    +0.4729 \\

    \midrule

    \raisebox{-0.5\height}{\includegraphics[width=0.10\linewidth]{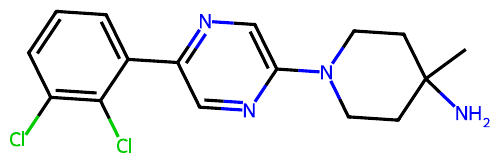}} &
    \raisebox{-0.5\height}{\includegraphics[width=0.10\linewidth]{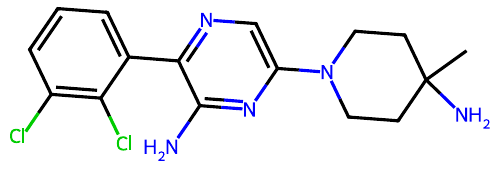}} &
    5EHR & 
    0.6667 & 
    -7.15\,/\, -9.75 & 
    -8.7560\,/\, -8.5590 &
    +0.8203\,/\, +0.7920 &
    +1.6715\,/\, +2.0898 &
    -2.60 &
    +0.1970 &               
    -0.0283 &
    +0.4183 \\

    \midrule

    \raisebox{-0.5\height}{\includegraphics[width=0.10\linewidth]{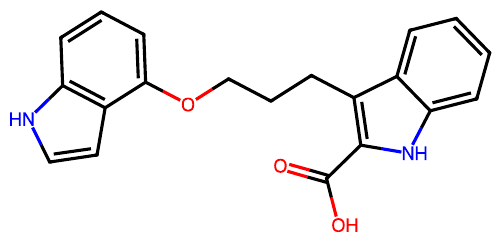}} &
    \raisebox{-0.5\height}{\includegraphics[width=0.10\linewidth]{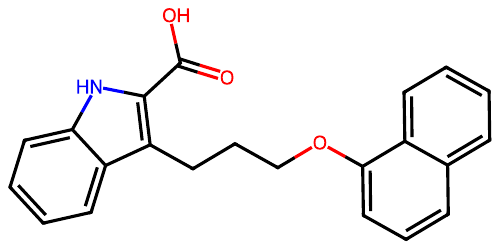}} &
    4HW3 & 
    0.7872 & 
    -6.66\,/\, -8.90 & 
    -3.5197\,/\, -7.2962 &
    +0.5356\,/\, +0.5435 &
    +1.5055\,/\, +1.7195 &
    -2.24 &
    -3.7765 &               
    +0.0078 &
    +0.2141 \\

    \midrule

    \raisebox{-0.5\height}{\includegraphics[width=0.10\linewidth]{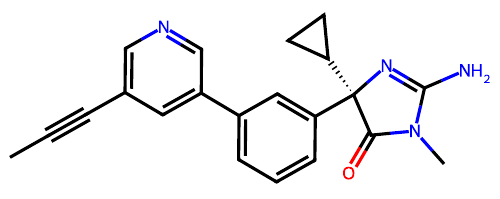}} &
    \raisebox{-0.5\height}{\includegraphics[width=0.10\linewidth]{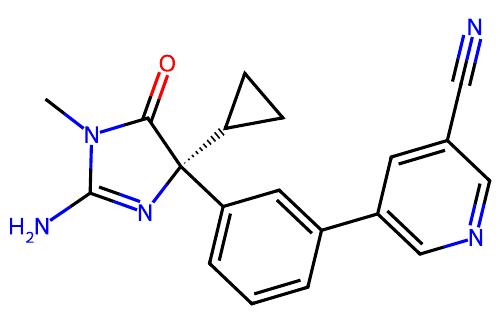}} &
    4DJW & 
    0.8103 & 
    -11.35\,/\, -9.42 & 
    -11.1010\,/\, -9.4110 &
    +0.9175\,/\, +0.9263 &
    +2.3914\,/\, +2.1834 &
    +1.93 &
    +1.6900 &               
    +0.0088 &
    -0.2080 \\

    \midrule

    \raisebox{-0.5\height}{\includegraphics[width=0.10\linewidth]{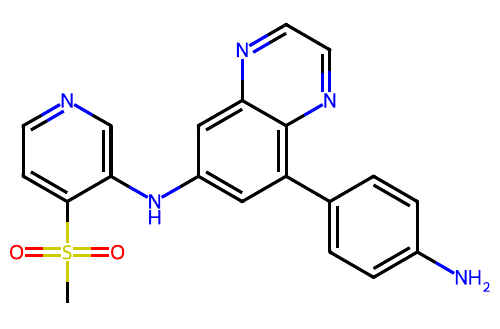}} &
    \raisebox{-0.5\height}{\includegraphics[width=0.10\linewidth]{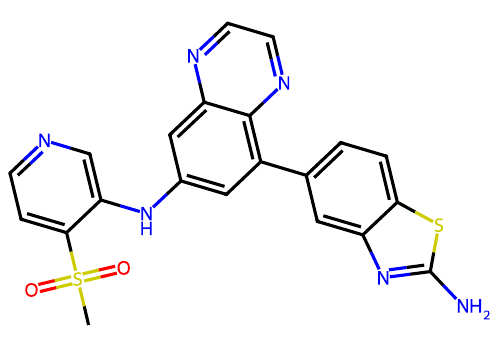}} &
    6HVI & 
    0.7458 & 
    -7.93\,/\, -10.24 & 
    -6.5430\,/\, -11.3100 &
    +0.6113\,/\, +0.6760 &
    +1.5235\,/\, +1.9901 &
    -2.31 &
    -4.7670 &               
    +0.0644 &
    +0.4666 \\

    \midrule

    \raisebox{-0.5\height}{\includegraphics[width=0.10\linewidth]{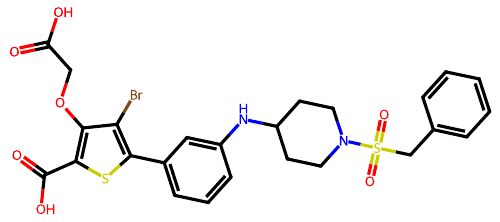}} &
    \raisebox{-0.5\height}{\includegraphics[width=0.10\linewidth]{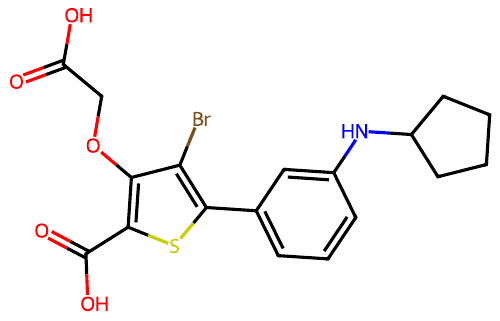}} &
    2QBS & 
    0.7385 & 
    -11.42\,/\, -8.72 & 
    -9.6890\,/\, -8.8168 &
    +0.9224\,/\, +0.8660 &
    +2.4139\,/\, +2.0398 &
    +2.70 &
    +0.8722 &               
    -0.0561 &
    -0.3741 \\

    \midrule

    \raisebox{-0.5\height}{\includegraphics[width=0.10\linewidth]{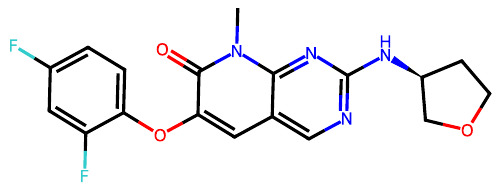}} &
    \raisebox{-0.5\height}{\includegraphics[width=0.10\linewidth]{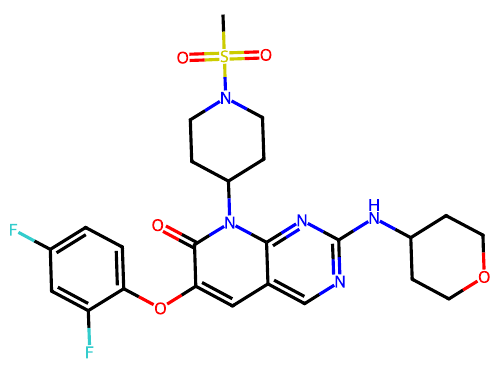}} &
    3FLY & 
    0.6351 & 
    -10.23\,/\, -12.26 & 
    -9.8951\,/\, -12.1479 &
    +0.2379\,/\, +0.2490 &
    +1.7866\,/\, +2.1387 &
    -2.03 &
    -2.2528 &               
    +0.0111 &
    +0.3522 \\

    \midrule

    \raisebox{-0.5\height}{\includegraphics[width=0.10\linewidth]{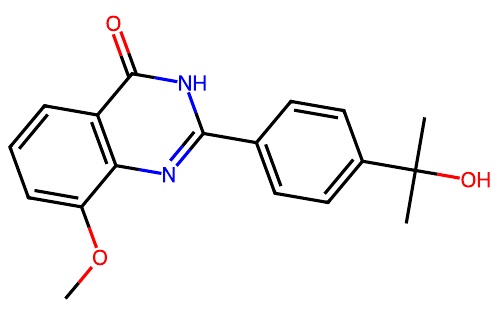}} &
    \raisebox{-0.5\height}{\includegraphics[width=0.10\linewidth]{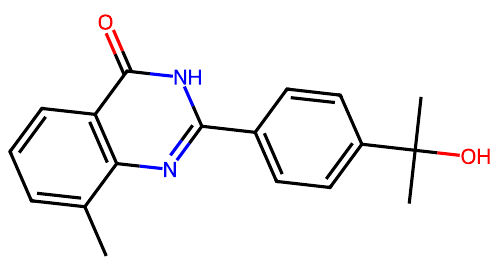}} &
    4UI5 & 
    0.7234 & 
    -10.05\,/\, -12.08 & 
    -9.7050\,/\, -11.9640 &
    +0.9090\,/\, +0.9380 &
    +1.9309\,/\, +2.2185 &
    -2.03 &
    -2.2590 &               
    +0.0288 &
    +0.2876 \\

    \midrule

    \raisebox{-0.5\height}{\includegraphics[width=0.10\linewidth]{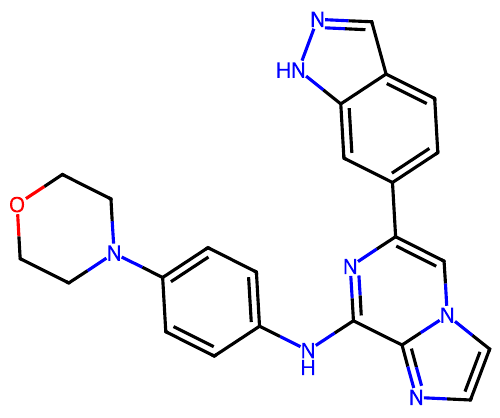}} &
    \raisebox{-0.5\height}{\includegraphics[width=0.10\linewidth]{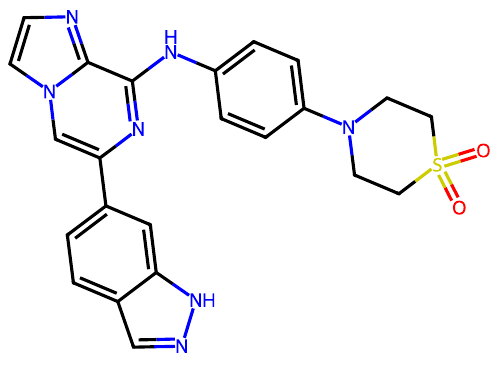}} &
    4PV0 & 
    0.8000 & 
    -6.82\,/\, -11.83 & 
    -10.7470\,/\, -11.1020 &
    +0.6660\,/\, +0.7170 &
    +1.9251\,/\, +2.1345 &
    -5.01 &
    -0.3550 &               
    +0.0508 &
    +0.2094 \\

    \midrule

    \raisebox{-0.5\height}{\includegraphics[width=0.10\linewidth]{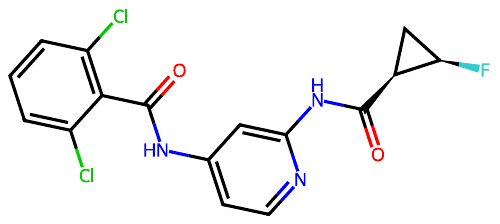}} &
    \raisebox{-0.5\height}{\includegraphics[width=0.10\linewidth]{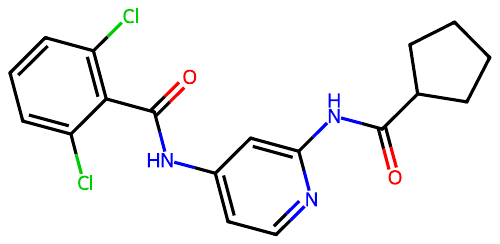}} &
    4GIH & 
    0.7347 & 
    -11.70\,/\, -9.00 & 
    -10.9067\,/\, -8.8033 &
    +0.7450\,/\, +0.7090 &
    +1.9382\,/\, +1.4873 &
    +2.70 &
    +2.1034 &               
    -0.0361 &
    -0.4510 \\

    \midrule

    \raisebox{-0.5\height}{\includegraphics[width=0.10\linewidth]{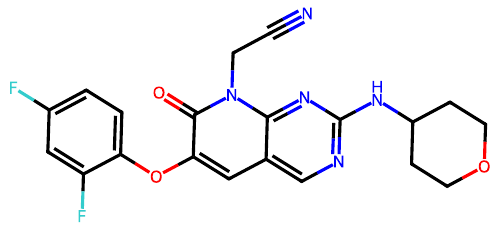}} &
    \raisebox{-0.5\height}{\includegraphics[width=0.10\linewidth]{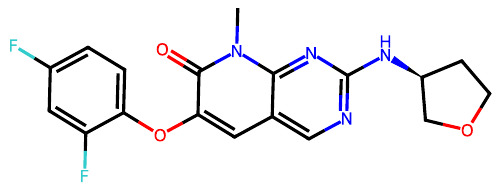}} &
    3FLY & 
    0.6571 & 
    -11.85\,/\, -10.23 & 
    -12.8311\,/\, -9.8951 &
    +0.2678\,/\, +0.2379 &
    +2.1754\,/\, +1.7866 &
    +1.62 &
    +2.9360 &               
    -0.0299 &
    -0.3889 \\

    \bottomrule

  \end{tabular}%
  }
\end{table*}

\subsection{Analysis of Cross-Target Activity–Cliff Pairs}
\label{sec:cliff-analysis}

Table~\ref{tab:cliff-examples} lists 21 ligand pairs whose ECFP\,\cite{doi:10.1021/ci100050t} similarity is greater than~0.60 yet display large differences in experimental binding free energy Exp\,$\Delta G$\, making them representative \emph{activity–cliff} cases for evaluating our embedding space. For comparison, the Euclidean scores in the table are produced by the current state-of-the-art pocket–ligand model LigUnity\(_{\mathrm{poc}}\)~\cite{feng2025foundation}, whereas the hyperbolic scores come from our method.

\noindent\textbf{Directional Agreement with Experimental Affinity.} Recall that a smaller (more negative) experimental \(\Delta G\) indicates a stronger binder, whereas a larger model score indicates stronger binding.  
Hence, for every pair in Table~\ref{tab:cliff-examples} we expect the sign of \(\Delta(\text{score})\) to be opposite to the sign of \(\Delta(\text{Exp}\,\Delta G)\).
This correspondence is clearly visible: whenever the experimental gap favours molecule B, the hyperbolic score is higher for B (positive \(\Delta\text{Hyp}\)), and vice-versa.
Euclidean scores occasionally match the sign but the margin is often negligible. Several pairs show that even free-energy perturbation (FEP)~\cite{wang2015accurate} predicts the wrong direction of the affinity change, yet the hyperbolic score still aligns with the experimental ordering.  

\noindent\textbf{Separation Magnitude.} The Euclidean score differences are typically tiny (many are \(<0.05\)), making it hard to tell the two ligands apart.  
In contrast, the hyperbolic score differences are an order of magnitude larger, providing an immediate visual cue of which ligand the model prefers.  
This numerical gap illustrates how the hyperbolic embedding stretches \emph{activity-cliff} pairs, whereas the Euclidean embedding leaves them almost collapsed.

\subsection{Evaluation Metrics}
\label{sec:eval-metrics}

Virtual screening asks whether a model can place a handful of true binders at the very top of a ranked list that may contain millions of inactives; affinity ranking asks whether it can preserve the fine-grained order of binding strengths within a chemically related series.  
Accordingly we employ different metrics.

\noindent\textbf{(1) Virtual Screening Metrics.}

\noindent\textbf{AUROC}.  
The area under the ROC curve is the probability that a randomly chosen active (\(a\)) scores higher than a randomly chosen inactive (\(d\)): \( \Pr[s(a)>s(d)]\).  
Values range from 0.5 (random) to~1.0 (perfect) but treat the whole ranked list uniformly.

\noindent\textbf{BEDROC\(_{80.5}\).}  
To emphasise the earliest part of the ranked list, we adopt the Boltzmann‐enhanced discrimination of ROC (BEDROC) with focus parameter \(\alpha = 80.5\), for which roughly the top 2 \% of ranks account for 80 \% of the score.  
Let \(N\) be the library size, \(N_{t}\) the number of actives, and \(r_i\in[1,N]\) the rank of active \(i\).  
The normalised form is
\begin{equation}
\label{eq:bedroc}
\begin{aligned}
\text{BEDROC}_{\alpha}
&=\;
\frac{\displaystyle
      \sum_{i=1}^{N_{t}} e^{-\alpha r_i / N}}
     {\displaystyle
      R_{\alpha}\!\left(
        \dfrac{1-e^{-\alpha}}{e^{\alpha/N}-1}
      \right)}
\\
&\quad
\times
\frac{R_{\alpha}\sinh(\alpha/2)}
     {\,\cosh(\alpha/2)-\cosh\!\bigl(\alpha/2-\alpha R_{\alpha}\bigr)}
\\
&\quad+\;
\frac{1}{1-e^{\alpha(1-R_{\alpha})}}\,,
\end{aligned}
\end{equation}
where \(R_{\alpha}=N_{t}/N\) is the active fraction.  
Equation~\eqref{eq:bedroc} is bounded in \([0,1]\); higher values indicate stronger early enrichment.

\noindent\textbf{Enrichment Factor}.  
The factor at a cut-off \(\alpha\%\) quantifies how many actives the model retrieves relative to random ranking:  
\begin{equation}\label{eq:ef}
  \text{EF}_{\alpha}
  =
  \frac{\text{NTB}_{\alpha}}
       {\text{NTB}_{t}\,\alpha/100},
\end{equation}
where \(\text{NTB}_{\alpha}\) is the number of true binders in the top \(\alpha\%\) of the list and \(\text{NTB}_{t}\) the total binders.

\noindent\textbf{ROC Enrichment (RE)}.  
At a false-positive-rate threshold \(x\%\) we report  
\begin{equation}\label{eq:re}
  \text{RE}(x\%) =
  \frac{\mathrm{TP}/P}{\mathrm{FP}_{x\%}/N}
  =
  \frac{\mathrm{TP}\,N}{P\,\mathrm{FP}_{x\%}},
\end{equation}
where \(N\) is the library size, \(P\) the number of actives, \(\mathrm{TP}\) the true positives among the top-ranked compounds, and \(\mathrm{FP}_{x\%}\) the false positives observed before the FPR reaches \(x\%\).  
A larger RE means stronger early discrimination.

\noindent\textbf{(2)Affinity-ranking metrics.}

Within a congeneric series we measure linear and rank agreement between predicted (\(\hat{y}\)) and experimental (\(y\)) affinities.

\begin{equation}\label{eq:pearson}
\text{Pearson } r \;=\;
\frac{\sum_{i}(y_i-\bar{y})(\hat{y}_i-\bar{\hat{y}})}
     {\sqrt{\sum_{i}(y_i-\bar{y})^{2}}\,
      \sqrt{\sum_{i}(\hat{y}_i-\bar{\hat{y}})^{2}}},
\end{equation}

\begin{equation}\label{eq:spearman}
\text{Spearman } \rho \;=\;
1-\frac{6\sum_{i}d_i^{2}}{n(n^{2}-1)},
\end{equation}
where \(d_i\) is the rank difference for compound \(i\) and \(n\) the series size.  
Both metrics lie in \([-1,1]\); higher values indicate better agreement (1 is perfect correlation).

\end{document}